# THE INDUCTIVE LOGIC OF INFORMATION SYSTEMS *

**Norman C. Dalkey**
Cognitive Systems Laboratory
Computer Science Department
UCLA, Los Angeles, CA. 90024-1600

**ABSTRACT**

An inductive logic can be formulated in which the elements are not propositions or probability distributions, but information systems. The logic is complete for information systems with binary hypotheses, i.e., it applies to all such systems. It is not complete for information systems with more than two hypotheses, but applies to a subset of such systems. The logic is inductive in that conclusions are more informative than premises. Inferences using the formalism have a strong justification in terms of the expected value of the derived information system.

## Information Systems

Information Systems (IS) are ubiquitous constructs in the knowledge sciences. Examples: surveillance systems, experiments, pattern recognition methods, coding schemes, expert systems, signal systems, sampling techniques... Despite the wide variety of incarnations, IS have a common underlying structure:

a.  A set $E$ of events of interest (hypotheses, states of nature, target events, etc.)

b.  A set $I$ of information events (observations, data, signals, messages, etc.)

c.  A joint probability distribution $P(EI)$ on combinations of hypotheses and observations.

In applications, the probabilities of interest are the posterior distributions $P(E \mid I)$ -- the conditional probabilities of hypotheses given observations. The role of the observations is to improve what is known about $E$.

To assess the value of an *IS* (e.g., to say precisely what is meant by *improve* in the preceeding sentence), it is necessary to have a figure of merit. In decision theory the figure of merit is the expected payoff of implementing the IS. In communication theory the figure of merit is information--an entropy measure. In pattern recognition, some form of error score is common. All such measures can be neatly summarized by the theory of proper scores.

A probabilistic score is a function $S(P, e)$ which assigns a rating (score, reward, payoff, etc.) to a probability distribution $P$ given that the event $e$ occurs. Such a score is called *proper* if it fulfills the condition:

$$\sum_E P(e) S(P,e) \geq \sum_E P(e) S(Q,e) \qquad (1)$$

that is, a score is proper if its expectation is a maximum when the distribution that determines

* This work was supported in part by National Science Foundation Grant IST 84-05161

the expectation is the same as the distribution which determines the score. Eq. (1) is roughly analogous to the requirement that an error score be a minimum when a response is correct.

The communication theory figure of merit stems from the logarithmic score, $S(P, e) = \log P(e)$. The expected score, $\sum_E P(e) \log P(e)$ is the negative of the entropy. Decision-theoretic scores are obtained from a decision matrix, $U(a, e)$, the payoff if an action $a$ is taken and the state of nature $e$ occurs. Let $a^*(P)$ be the optimal action if $P$ is the distribution on $E$. $S(P, e) = U(a^*(P), e)$ is a proper score. An analogue of least squared error is given by the quadratic score, $S(P, e) = 2P - \sum P(e)^2$. Abbreviate $\sum_E P(e) S(P, e)$ by $G(P)$ and $\sum_E P(e) S(Q, e)$ by $G(P, Q)$. Eq. (1) can be written as $G(P) \geq G(P, Q)$. It is clear that $G(P, Q)$ is linear in $P$, and it is relatively straightforward to show that $G(P)$ is convex in $P$. A basic result is (Proofs of theorems are in the Appendix):

**Theorem 1:**
> If $R = aP + (1-a)Q$, $0 \leq a \leq 1-a$, and $G(R) \geq G(Q)$ for all $a$, and $G(P, R)$ is continuous in $R$ at $Q$, then $G(P, Q) \geq G(Q)$.

The import of Theorem 1 is extended by its corollary: If $K$ is a convex set of probability distributions, and $Q = \arg\min_K H(P)$, then $G(P, Q) \geq G(Q)$ for all $P$ in $K$.

As an example of the application of the corollary, if all that is known about a probability distribution $P$ is that it is in a convex set $K$, and there is a minimum at $Q$ for the expected score $G(P)$ in $K$, then if $Q$ is taken as an estimate of the unknown $P$, the actual expectation $G(P, Q) \geq G(Q)$. In short, the expected observed score of $Q$ is guaranteed to be at least $G(Q)$. This feature of proper scores was used as the basis of a weaker theory of inductive logic on probability distributions in an earlier effort. [Dalkey, 1985].

The initial probability of an observation $i$ is $P(i) = \sum_E P(e.i)$. The expected value of an IS is $H(P) = \sum_I P(i) G(P(E \mid i), e)$. That is, the value of an information system is the average over the potential observations of the expected score of the posteriors. $H(P)$ for IS corresponds to $G(P)$ for probability distributions. The analogue for IS of $G(P, Q)$ is $H(P, Q) = \sum_I P(i) G(P(E \mid i), Q(E \mid i))$, the average over potential observations of the relative score of the posteriors of $P$ and $Q$. It measures the expected value if $Q$ is the estimated IS and $P$ is the actual IS.

It follows from the linearity of $G(P,Q)$ in $P$ and the convexity of $G(P)$, that the averages $H(P,Q)$ and $H(Q)$ have the same properties. It is also relatively straightforward to show that Theorem 1 and its corollary hold for IS.

## Dominance

A major advantage of IS over probability distributions for the representation of uncertainty is the fact that IS allow dominance. Given any two different probability distributions $P$ and $Q$, there is a score rule $S$ such that $G_S(P) > G_S(Q)$ and another score rule $S'$ such that $G_{S'}(P) < G_{S'}(Q)$. However, there can be two different IS $P$ and $Q$, such that $H(P) \geq H(Q)$ for all score rules. Define $P \geq Q$ to mean $H(P) \geq H(Q)$ for all proper scores.

It is clear that $\geq$ is a partial order, that is, it fulfills:

a. Transitivity: $P \geq Q$ and $Q \geq R \rightarrow P \geq R$.

b. Reflexivity: $P \geq P$

c. Antisymmetry: $P \geq Q$ and $Q \geq P \rightarrow P = Q$

Theorem 1 and its corollary carry over to dominance. For the corollary, if $K$ is a convex set of IS, and there is a $Q$ in $K$ such that $P \geq Q$ for every $P$ in $K$, then $H(P,Q) \geq H(Q)$ for all scores.

The partial order $\geq$ has a unique absolute upper bound, $P^*$ the IS such that for each member of $E$, there is an $i$, $P(e \mid i) = 1$. $P^*$ is often called *perfect information* in decision theoretic analyses of information. $\geq$ also has a unique absolute lower bound $P^0$, the IS $P^0(e.i) = P(e)P(i)$, or in other words, the prior probability $P(e)$ is taken to be the posterior for every observation $i$. It is straightforward to show that $P^* \geq P \geq P^0$ for every $P$. The statement $P \geq P^0$ is equivalent to the well-known positive value of information principle--information is never harmful providing it is free. [LaValle, 1978].

The dominance relation thus imposes a fair amount of structure on information systems. However, it is not quite enough to establish a logic.

## Logic

A partial order is called a lattice if, for each pair of elements $P$, $Q$, there is a unique least upper bound (*l.u.b.*) with respect to the ordering relation, i.e., there is an element $R$ such that $R \geq P$ and $R \geq Q$, and for any $R'$ which also dominates both, $R' \geq R$; and conversely, there is a greatest lower bound (*g.l.b.*) defined by replacing $\geq$ with $\leq$.

Lattices have received a great deal of attention as foundations for logics [Birkhoff, 1948]. Traditional two-valued logic can be represented by the Boolean lattice of sets. Other, more exotic logics have been formulated using different elements and different ordering relations.

The relation $\geq$ is not, in general a lattice. Examples exist of pairs of IS for which there is no unique $l.u.b.$ [Dalkey, 1980]. For one significant class of IS, however, dominance is a lattice, namely the set of IS with binary hypotheses, those where the events of interest are yes-no type: Will the patient die? Is the enemy intending an attact? Will a Democrat be elected president in 1988?

**Theorem 2:**
>For any pair of IS $P, Q$ with binary hypotheses, there is an IS, $P + Q$, which is the $l.u.b.$ of $P$ and $Q$ with respect to $\geq$.

$P + Q$ can be thought of as the minimal composition of $P$ and $Q$; i.e., it is the least informative of all IS which incorporate the information in both $P$ and $Q$.

Designate the $g.l.b.$ of $P$ and $Q$ by $P \cdot Q$. $P \cdot Q$ represents the IS which contains just the information which is common to both $P$ and $Q$.

The lattice $\geq$ on IS with binary hypotheses is not isomorphic to the Boolean lattice of two-valued logic. In particular, it does not have a negation (complement) nor is it distributive, i.e., $P \cdot (Q + R) \neq P \cdot Q + P \cdot R$. Thus, it affords a somewhat less powerful calculus than Boolean logic. However, a number of analogies exist. Thus we have:

$$P^* + P = P^* \qquad\qquad P^0 + P = P$$

$$P^* \cdot P = P \qquad\qquad P^0 \cdot P = P^0$$

Other familiar properties: Idempotence, $P+P = P, P \cdot P = P$. Commutativity, $P+Q = Q+P, P \cdot Q = Q \cdot P$. Associativity, $P+(Q+R) = (P+Q)+R, P \cdot (Q \cdot R) = (P \cdot Q) \cdot R$. Consistency, $P \geq Q, P+Q = P, P \cdot Q = Q$ are equivalent. Absorption, $P+(P \cdot Q) = P \cdot (P+Q) = P$. Semi-distributivity, $P+(Q \cdot R) \leq (P+Q) \cdot (P+R)$, $P \cdot (Q+R) \geq (P \cdot Q) + (P \cdot R)$. Semi-cancellation, $P+Q = P+R$ and $P \cdot Q = P \cdot R$ implies $Q = R$.

Thus, anyone familiar with Boolean logic will find that many of the transformations carry over to IS logic.

**Inference**

In classical logic, the basic characteristic of an inference can be stated as: If the premises are true, then the conclusion must be true. This stark precept cannot be maintained for *IS* logic. The basic inference schema in *IS* logic is just: if $P$ and $Q$ are known, then assert $P + Q$. But there is no justification for believing that $P + Q$ is the actual composition of $P$ and $Q$. Instead, there is a strong a-fortiori argument for selecting $P + Q$ in the absence of knowing the actual composition.

If $P$ and $Q$ are known, but the true composition $R$ is not known, then there is a set $K$ of compositions compatible with both $P$ and $Q$, and $R$ could be any member of $K$. $P + Q$ must be a member of $K$ (see Appendix) and since $P + Q$ is the l.u.b. of $P$ and $Q$, for any other member $R$ of $K$, $R \geq P + Q$. On the other hand for the same reason, $P + Q \geq P$ and $P + Q \geq Q$. From the extension of the corollary to Theorem 1 to $IS$ and $\geq$, we can assert that $H(R, P + Q) \geq H(P + Q) \geq \max [H(P), H(Q)]$ for all score rules. In short, the expected value of $P + Q$ is at least as great as what it "promises", and is at least as great as the better of $P$ or $Q$. In most cases, of course, the cautious "greater than or equal" will turn out to be a more optimistic "greater".

In a more colloquial vein, if a doctor is concerned about a specific disease for a particular patient, and he knows two different tests for the disease, but does not know the correlation between the tests, then, whether he is interested primarily in money, fame, or the welfare of the patient, he is better off to use the minimal composition than to use either test alone.

To recap to this point: For $IS$ with binary hypotheses, a complete logic exists which is analogous to, but not identical with, the traditional two-valued logic. Inferences carry with them the strong guarantee that the actual expected value of conclusions is at least as great as the computed expected value for all score rules (i.e., for all payoff functions.) It is this feature which justifies the use of the term *logic*.

### Non-binary Hypotheses

Although $IS$ with binary hypotheses are significant in practical affairs, the majority of $IS$ of direct interest to AI have a multiplicity of hypotheses. As noted earlier, it is not difficult to find examples of pairs of $IS$ with multiple hypotheses that have a l.u.b. with respect to $\geq$. For such pairs, deriving $P + Q$ when $P$ and $Q$ are known carries all the positive features attending a similar inference with $IS$ with binary hypotheses.

It clearly would be desirable to have operational criteria to distinguish pairs of multiple hypothesis $IS$ that have a l.u.b. with respect to $\geq$. In particular, this would allow identifying cases of incomplete information in which no ad-hoc assumptions such as independence are necessary to fill in the missing information. At present, only weak necessary and non-operational sufficient conditions are known (see Appendix).

Even with operational criteria for l.u.b.'s, it seems likely that many cases of $IS$ of interest to AI will not fit the criteria. For such cases, the question arises whether $IS$ logic is relevant. One clear application is: For any pair of $IS$ $P$ and $Q$, a set $D$ can be defined which consist of the $IS$ that dominate both $P$ and $Q$ but which do not dominate any $R$ in the set $K$ of $IS$ that dominate $P$ and $Q$. $D$ can play the same role for $IS$ logic that the Pareto-optimal set plays in economics and decision theory -- i.e., to rule out clearly inappropriate conclusions.

A fallback tactic is available which is not entirely dismal. If a specific score rule is adopted then there is a complete order on all *IS*. For any pair of *IS* a l.u.b. exists except for epsilontics where the relative score is not always continuous in its second argument. (One plus for the logarithmic score is that $H(P,Q)$ is always continuous in $Q$ providing it is in the interior of the space of *IS*.) For the given score, the inference from $P$ and $Q$ to $P+Q$ carries its own guarantee -- $H(R, P+Q) \geq H(P+Q)$ for all $R$ compatible with both $P$ and $Q$. For example, interpreting the expected logarithmic score as a measure of the amount of information in an *IS*, the amount of information in $P+Q$ is always greater than the information in either $P$ or $Q$. Thus $P+Q$, even if based on a single score rule, is more "solid" than, say, tactical assumptions of independence.

In sum, for *IS* with binary hypotheses a complete inductive logic exists. Only pieces of a logic exist for *IS* with non-binary hypothesis, but they can be valuable to AI research. Finally, clearly a number of live research topics remain.

# APPENDIX

*Theorem 1.* If $R = aP + (1-a)Q$, $0 \leq a \leq 1$, $G(R) \geq G(Q)$ for all $a$, and $G(P, R)$ is continuous in $R$ at $Q$, then $G(P, Q) \geq G(Q)$.

*Proof:* $G(R) = \sum_E (aP(e) + (1-a)Q(e))S(R, e) = aG(P, R) + (1-a)G(Q, R)$.
Since $G(Q) \geq G(Q, R)$ and $G(R) \geq G(Q)$, $G(P, R) \geq G(Q)$. If $G(P, R)$ is continuous in $R$ at $Q$, $G(P, R) \geq G(Q)$.

If $G(P, R)$ is not continuous at $Q$, the only troublesome case is that in which $G(R, Q)$ has negative slope (with increasing $a$). This can only occur if $G(P)$ has an absolute minimum at $Q$. If mixed strategies are introduced, the score rule $S$ can be replaced by $S'$, where $G'(R, Q) = G(Q)$, all $R$. Call an $S'$ so modified *regular*. Theorem 1 is true for all regular score rules, without the continuity restriction. It will be assumed that only regular score rules are being addressed in the following.

*Corollary 1.* If $K$ is a convex set of probability distributions, and $Q = arg \min_K G(P)$, then $G(P, Q) \geq G(Q)$ for all $P$ in $K$.

*Proof:* Since $K$ is convex, $R = aP + (1-a)Q$ is in $K$, and by the assumption of minimality of $Q$, $G(R) \geq G(Q)$. Thus the hypotheses of Theorem 1 are fulfilled and $G(P, Q) \geq G(Q)$.

To prepare for Theorem 2, let $t(i)$ be the vector $(P(i|e), P(i|\bar{e}))$, and arrange these vectors in decreasing order of $P(i|e)/P(i|\bar{e})$. For simplicity, relabel the observations in I with the given order -- i.e., $t(1)$ is the $t(i)$ with the highest slope. Let $T(i) = \sum_{j \leq i} t(j)$. If these vectors are plotted in the unit square, they determine a concave, piece-wise linear curve that lies above or on the diagonal. $T(0) = (0, 0)$. Let $m$ be the number of observations; $T(m) = (1, 1)$, since $\sum_I P(i \mid e) = 1$. Let $C(P)$ be the convex closure of the points in this curve. I call $C(P)$ the canonical representation of $P$. $C(P)$ contains the diagonal, which is $C(P^\circ)$, the canonical representation of the prior information system.

*Theorem 2.* If $P$ and $Q$ are binary-hypothesis $IS$, the l.u.b., $P + Q$, exists.

*Proof:* Set $P+Q = <(C(P), C(Q)>$, the convex closure of $C(P)$ and $C(Q)$. $C(P) \subset P+Q$, and $C(Q) \subset P+Q$. Thus, $P+Q \geq P$ and $P+Q \geq Q$ [Dalkey, 1980, Theorem 7]. Thus, $P + Q$ is an $IS$ which is an upper bound of $P$ and $Q$. If there were another upper bound $R$ such that $P + Q > R$, $C(R) \subset C(P + Q)$, but also $C(P) \subset C(R)$ and $C(Q) \subset C(R)$. Since the $C$'s are convex, no such $R$ exists.

The canonical representation is a simple, constructive method of computing $P + Q$ for any pair of $IS$ with binary hypotheses. For $IS$ with moderate numbers of $I$'s, the construction is easy to do by hand. For more complex $IS$, the construction is easily programmable for a computer.

Canonical representations can be defined for $IS$ with more than two hypotheses. A sufficient condition for $P + Q$ to exist for such $IS$ is that the convex closure $< C(P), C(Q) >$ be itself an $IS$. However, at present, that condition cannot be expressed in algorithmic form.